\documentclass[runningheads]{llncs}

\usepackage{eccv}

\usepackage{eccvabbrv}

\usepackage{graphicx}
\usepackage{booktabs}
\usepackage{multirow}

\usepackage[accsupp]{axessibility}  

\usepackage{hyperref}

\usepackage{orcidlink}

\begin{document}

\title{OmniDance: Multimodal Driven Dance Video Generation with Large-scale Internet Data} 

\titlerunning{OmniDance}

\newcommand{\projectlead}{\textsuperscript{$\ddagger$}}
\newcommand{\corrauth}{\textsuperscript{\textdagger}}

\author{
Kaixing Yang\inst{1}\orcidlink{0000-0002-3879-7225} \and
Jiashu Zhu\inst{2}\projectlead\orcidlink{0009-0005-0289-9704} \and
Xulong Tang\inst{5}\orcidlink{0009-0005-0857-1056} \and
Ziqiao Peng\inst{1}\orcidlink{0009-0000-1341-9618} \and
Xiangyue Zhang\inst{4}\orcidlink{0009-0002-5642-9474} \and
Chubin Chen\inst{3}\orcidlink{0009-0004-4169-5491} \and
Puwei Wang\inst{1}\corrauth\orcidlink{0000-0002-2949-5380} \and
Jiahong Wu\inst{2}\corrauth\orcidlink{0000-0001-8583-0414} \and
Xiangxiang Chu\inst{2}\orcidlink{0000-0003-2548-0605} \and
Hongyan Liu\inst{3}\corrauth\orcidlink{0000-0002-4902-1078} \and
Jun He\inst{1}\corrauth\orcidlink{0000-0003-1511-7554}
}

\authorrunning{K.~Yang et al.}

\institute{
Renmin University of China, Beijing, China\\
\email{\{yangkaixing,pengziqiao,wangpuwei,hejun\}@ruc.edu.cn}
\and
AMAP, Alibaba Group, Beijing, China\\
\email{\{zhujiashu.zjs,hongxi.wjh\}@alibaba-inc.com, cxxgtxy@gmail.com}
\and
Tsinghua University, Beijing, China\\
\email{liuhy@sem.tsinghua.edu.cn, trubeeen@gmail.com}
\and
Wuhan University, Wuhan, China\\
\email{xiangyuezhang@whu.edu.cn}
\and
Malou Tech Inc, Plano, Texas, USA\\
\email{xulong.tang@maloutech.com}
}

\maketitle

\begingroup
\renewcommand{\thefootnote}{}
\footnotetext{\textsuperscript{$\ddagger$}Project leader.}
\footnotetext{\textsuperscript{\textdagger}Corresponding authors.}
\endgroup

\begin{abstract}
Music-driven dance video generation aims to synthesize expressive human motion that is temporally aligned with music while maintaining high visual fidelity.
Despite recent progress, existing methods still struggle to generate dance videos that simultaneously exhibit expressive motion and high visual quality.
This limitation primarily arises from two factors:
\textit{(1) Dataset.} The lack of large-scale and high-quality datasets and effective data collection pipelines specifically tailored for dance video generation; and
\textit{(2) Method.} The absence of principled framework-level solutions for effectively integrating music as a complementary conditioning signal into the Video Generation Foundation Models.
To address the dataset limitation, we introduce \textbf{CIPE-Dance}, a large-scale Internet-sourced dance video dataset, equipped with \textbf{C}horeograph \textbf{I}nformed text annotations and constructed via a \textbf{P}rogressive \textbf{E}xpert pipeline. To the best of our knowledge, \textbf{CIPE-Dance} is the largest dataset for dance video generation to date, comprising 300k high-quality clips (over 400 hours) and covering diverse dancers, environments, and dance genres.
To overcome the method limitation, we propose \textbf{OmniDance}, a framework-level recipe for integrating music into a TI2V foundation model without sacrificing its original controllability or visual fidelity.  Motivated by the complementary roles of text (low-frequency semantics) and music (high-frequency temporal dynamics), OmniDance co-designs a depth-aware specialization model architecture, an anchored easy-to-hard curriculum learning strategy, and modality-specialized time-dependent CFG strategy, achieving unified TI2V/MI2V/MTI2V generation.
Extensive experiments on the \textbf{CIPE-Dance} dataset demonstrate that \textbf{OmniDance} achieves state-of-the-art performance across TI2V, MI2V, and MTI2V tasks, while exhibiting robust multimodal integration capability. 
Project is available at https://github.com/AMAP-ML/OmniDance.
\end{abstract}

\begin{figure}[t]
  \centering
  \includegraphics[width=\linewidth]{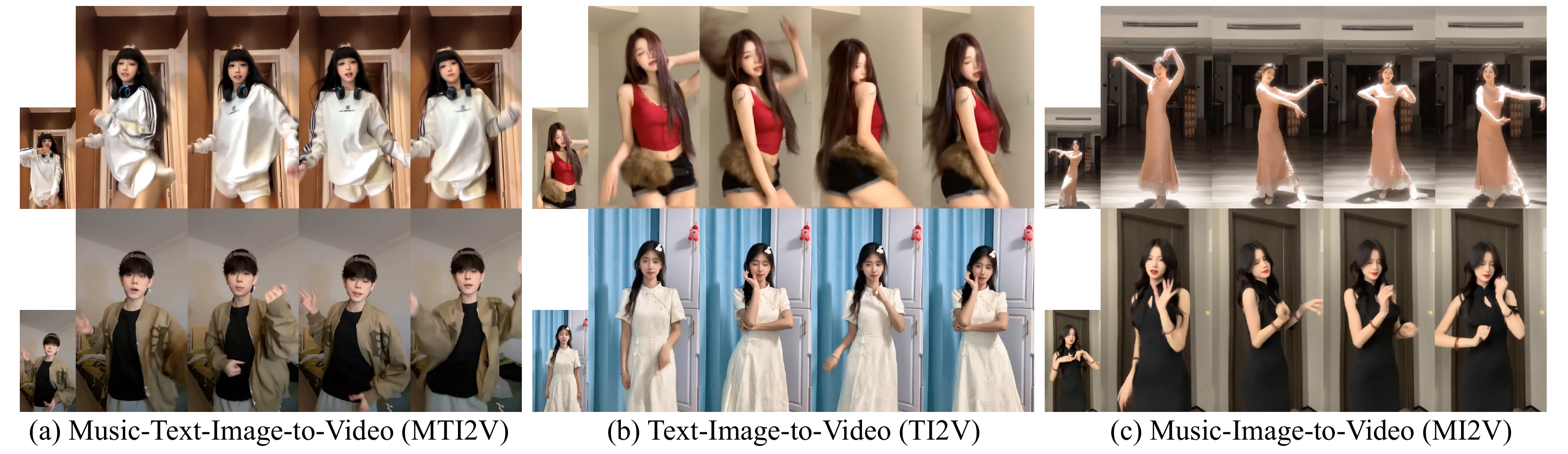}
  \caption{Given a reference image, OmniDance supports multimodal driven dance video generation, including music only (MI2V, right), text only (TI2V, middle), and both (MTI2V, left). OmniDance not only preserves identity and visual fidelity, but also produces artistically expressive and music-aligned motion.}
  \label{fig:teaser}
\end{figure}

\section{Introduction}
Dance is a vital part of human culture: through choreographed movement, dancers communicate emotion and narrative intent while showcasing the beauty of human motion~\cite{yang2025matchdance,butterworth2004teaching}. 
In the internet era, dance videos have become a dominant form of visual content on platforms such as YouTube and TikTok, and recent advances in AI-generated content (AIGC) have made high-fidelity video synthesis increasingly accessible~\cite{chen2025finger,li2025ld,feng2025narrlv,lei2025there,ling2025vmbench,chen2025s,chen2026taming,hu2026embedding}. 
Crucially, dance is jointly governed by \emph{semantic intent} (e.g., style and choreographic goals) and \emph{musical structure} (e.g., rhythm, tempo, and energy evolution)~\cite{yang2024beatdance}, motivating multimodal-driven dance video generation under text, music, and their combination.

In recent years, substantial progress has been made in several dance video related research directions, including talking head generation~\cite{peng2024synctalk,peng2025actavatar,peng2025omnisync}, pose-driven image animation~\cite{hu2024animate,tan2024animate}, and music-driven 3D dance generation~\cite{tseng2023edge,siyao2022bailando}. 
However, talking head generation does not model the tight correspondence between audio and full-body motion~\cite{cui2024hallo2,peng2025synctalk++}, pose-driven image animation overlooks the central role of music in dance composition~\cite{xu2024magicanimate,gao2025wan}s, and music-driven 3D dance generation is typically trained on small-scale datasets, ultimately constraining their motion generation capacity and generalization to diverse in-the-wild settings~\cite{yang2025megadance,yang2025flowerdance,jia2026bitdiff}. 
Despite growing interest, progress in music-driven dance video generation remains limited. 
Early works~\cite{chen2025x,park2025m2pe,tang2025spatial} first predict 2D keypoints from music and subsequently drive image-based animation, while others~\cite{wang2025choreomuse,yang2025mace} instead infer 3D SMPL parameters and render image animations accordingly. 
Moreover, ~\cite{wang2025dance,dong2025every} introduce end-to-end diffusion-based generation pipelines. 
However, existing methods still struggle to generate dance videos that simultaneously exhibit expressive motion and high visual quality, primarily due to two factors.
\textit{(1) Dataset.} The lack of large-scale open datasets and effective data collection pipelines specifically tailored for dance video generation.
\textit{(2) Method.} The absence of principled framework-level solutions for effectively integrating music as a complementary conditioning signal into Video Generation Foundation Models, which are trained without audio supervision.

To tackle the dataset limitation, we firstly develop a \textit{Progressive Expert-Based Data Collection Pipeline}.
To improve computational efficiency, we adopt a progressive easy-to-hard strategy.
For simple task, a lightweight verification expert (Qwen3-VL-2B~\cite{bai2025qwen3vltechnicalreport}) is sufficient to directly predict positive cases.
For more complex scenarios, we employ multiple heavier filtering experts (Qwen3-VL-8B~\cite{bai2025qwen3vltechnicalreport}) to inversely identify and eliminate common failure modes (i.e., negative cases). Specifically, the pipeline consists of the following six stages: (1) Popular Creator Mining, (2) Visual quality verification, (3) Reference clarity verification, (4) Dance video verification, (5) Single-dancer filtering, and (6) Scene stability filtering.
Secondly, we adopt \textit{Choreograph Informed Text Annotations}, under which each video is annotated from five complementary choreography-informed aspects using Qwen3-VL-8B~\cite{bai2025qwen3vltechnicalreport}:
(1) Body Dynamics,
(2) Choreographic Content,
(3) Expressiveness,
(4) Camera Presentation,
(5) Overall Look.
Finally, we propose \textbf{CIPE-Dance}, the large-scale Internet-source dataset, equipped with the \textbf{C}horeography \textbf{I}nformed annotations and constructed by the \textbf{P}rogressive \textbf{E}xpert pipeline. To the best of our knowledge, the \textbf{CIPE-Dance} dataset is currently the largest open-source dataset for dance video generation, comprising approximately 300k clips (over 400 hours) and covering diverse dancer demographics, performance environments, and over 30 dance genres. 

To address the methodological limitation, we propose \textbf{OmniDance}, a unified framework that supports multimodal driven dance video generation (TI2V, MI2V, and MTI2V), as shown in Fig. \ref{fig:teaser}. \textbf{OmniDance} effectively injects music conditioning into a Video Generation Foundation Models, without compromising its original capabilities, through a three-level design.
\textbf{(1) Model Architecture.}
Built upon WAN2.2-TI2V-5B~\cite{wan2025wanopenadvancedlargescale}, we augment each DiT block with an additional \emph{music} cross-attention layer. 
Motivated by the depth hierarchy of DiT models—where shallow layers capture low-frequency global structure and deeper layers refine high-frequency details—we design a \textit{Music-Text Progressive Specialization} scheme: shallow layers emphasize text conditioning to establish semantic layout, while deeper layers progressively strengthen music conditioning to refine rhythm-aware motion dynamics via the audio residual.
\textbf{(2) Training Strategy.}
We adopt an \textit{Easy-to-Hard Curriculum Learning Strategy} to extend the pretrained TI2V model toward unified TI2V, MI2V, and MTI2V generation \emph{while preserving} its original text/image controllability and visual fidelity.
In Stage I, \textbf{OmniDance} is trained exclusively on TI2V to adapt the model to dance-specific textual descriptions.
In Stage II, a music branch is introduced, and \textbf{OmniDance} jointly performs TI2V and MTI2V, enabling music understanding under text guidance.
In Stage III, \textbf{OmniDance} is trained on all three tasks—TI2V, MI2V, and MTI2V—allowing the model to generate dance videos solely from music when text is absent.
\textbf{(3) Inference Strategy.}
For classifier-free guidance (CFG), we propose a \textit{Modality-Specialized CFG} that progressively \emph{shifts the guidance emphasis} from text to music along the sampling trajectory: text dominates early steps to establish global semantics, while music becomes increasingly influential at later steps to refine rhythm-aware motion dynamics.

In conclusion, our contributions are as follows:
(1) We introduce \textbf{CIPE-Dance}, the largest dance video generation dataset to date, equipped with choreography informed text annotations and constructed via a progressive expert pipeline.
(2) We propose \textbf{OmniDance}, a three-level co-designed framework (architecture, curriculum learning, and modality-specialized inference) for injecting music conditioning into TI2V foundation models while preserving controllability and visual fidelity.
(3) We unify \textbf{TI2V}, \textbf{MI2V}, and \textbf{MTI2V} in a single model (\textbf{OmniDance}), enabling seamless modality switching at inference time without training or maintaining separate generators.
(4) We conduct extensive experiments on the \textbf{CIPE-Dance} dataset, demonstrating the state-of-the-art (SOTA) performance of \textbf{OmniDance} on TI2V, MI2V, and MTI2V generation, together with its robust multi-modal integration capability.

\section{Related Work}
\subsection{Human Motion Generation}
Human motion generation has advanced rapidly in recent years and is closely related to dance video generation, particularly in pose-driven image animation, talking head generation, and music-driven 3D dance generation.
\textbf{(1) Pose-driven image animation.} Pose-driven image animation utilizes 2D keypoints to generate motion videos, achieving notable advances~\cite{tan2024animate,hu2024animate,cheng2025wan}. However, these methods overlook the fact that music constitutes the structural backbone of dance. Generating dance motions by choreographing pose sequences from music—rather than animating pre-specified poses—represents a substantially more valuable and challenging problem.
\textbf{(2) Talking head generation.} Talking head generation employs audio features to generate vivid and lip-synced videos, also achieving significant breakthroughs~\cite{peng2024synctalk,peng2025synctalk++,peng2025omnisync}. 
However, these approaches fail to establish a principled correspondence between audio signals and full-body motion, even though such alignment is essential for the aesthetic expression of dance.
\textbf{(3) Music-driven 3D dance generation.} 
Music-driven 3D dance generation aims to synthesize full-body motion from music~\cite{yang2024codancers,yang2026tokendance,gong2023tm2d,yang2024cohedancers,li2023finedance}.
Beyond dance-specific studies, speech-driven 3D gesture generation has also made notable progress in recent years~\cite{zhang2025semtalk,zhang2025echomask,zhang2026mitigating,zhang2026personagesture,zhou2026not,zhang2025robust}.
However, most existing 3D dance methods are developed on relatively small-scale datasets (e.g., FineDance—the largest publicly available 3D dance dataset to date—contains only approximately 8 hours of motion), which ultimately constrains their motion generation capacity and generalization to diverse in-the-wild settings.

\subsection{Music-Driven Dance Video Generation}
Finally, research on music-driven dance video generation remains limited and can be broadly categorized into three paradigms: 2D keypoint–based generation, 3D SMPL-based generation, and end-to-end video generation.
\textbf{(1) 2D keypoint-based generation.} 
X-Dancer~\cite{chen2025x}, M2PE-Diff~\cite{park2025m2pe}, and STG-Mamba~\cite{tang2025spatial} first predict 2D keypoints from music and then drive image-based animation, but they remain challenged by limb occlusions and complex full-body locomotion in dance videos.
\textbf{(2) 3D SMPL-based generation.} 
ChoreoMuse~\cite{wang2025choreomuse} and MACE-Dance~\cite{yang2025mace} instead infer 3D SMPL parameters and render image animation accordingly. However, the high cost of acquiring 3D supervision keeps existing 3D dance datasets small in scale, limiting the diversity and fidelity of the learned motion patterns.
\textbf{(3) End-to-end generation.}
DabFusion~\cite{wang2025dance}, MusicInfuser~\cite{Hong_2026_CVPR} and MuseDance~\cite{dong2025every} introduce end-to-end generation pipelines; however, they do not effectively leverage Video Generation Foundation Models, resulting in in-the-wild outputs with limited motion expressiveness and degraded visual appearance.

In conclusion, existing methods still struggle to generate dance videos that simultaneously exhibit expressive motion and high visual quality.
This limitation primarily arises from two factors:
\textit{(1) Dataset.} The lack of large-scale and high-quality datasets and effective data collection pipelines specifically tailored for dance video generation; and
\textit{(2) Method.} The absence of principled framework-level solutions for effectively integrating music as a complementary conditioning signal into Video Generation Foundation Models, which are trained without audio supervision.

\begin{figure*}[t]
  \centering
  \includegraphics[width=\linewidth]{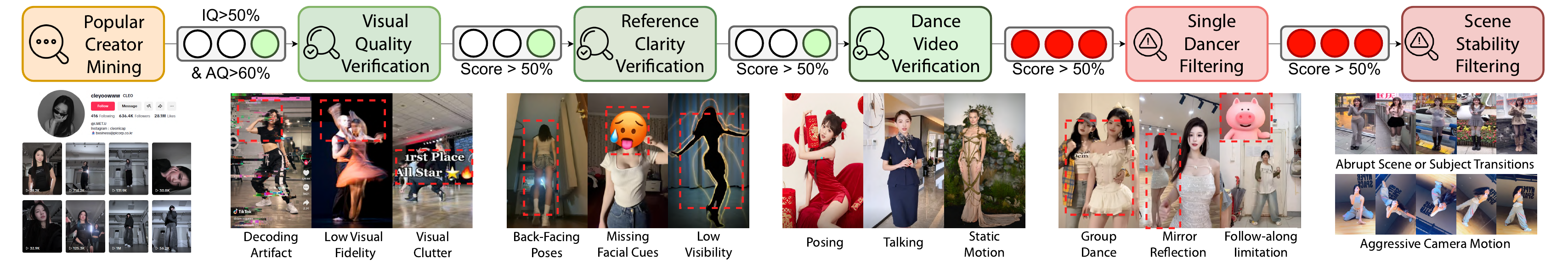}
  \caption{Overview of the Progressive Expert-Based Data Collection Pipeline.}
  \label{fig:data}   
\end{figure*}

\section{Dataset}
\subsection{Data Collection}
To address data quality challenges in large-scale web-crawled dance video collection, we develop a \textit{Progressive Expert-Based Data Collection Pipeline}, as shown in Fig. ~\ref{fig:data}. 
To improve computational efficiency, the pipeline adopts a progressive easy-to-hard data filtering strategy.
For simple cases, a lightweight verification expert is sufficient to directly identify positive samples. 
For hard scenarios, we employ multiple moderately heavier filter experts to inversely identify and eliminate common failure modes (i.e., negative samples).  
Specifically, the pipeline consists of the following stages.
\noindent\textbf{(1) Popular Creator Mining.} 
We collect all publicly available videos from approximately 500 dance content creators on Douyin or TikTok. Creators are required to have more than 50K followers, serving as a proxy for content quality and production standards. To ensure popularity, we restrict the dataset to videos published after 2018. 
The collected videos are segmented into 5-second clips at 16 FPS for downstream processing, resulting in a total of 630k clips.
\noindent\textbf{(2) Visual quality verification.} 
To ensure visual fidelity, we apply VBench~\cite{li2024mvbench} for quality filtering. 
We retain only clips with image quality (IQ) scores above 60.00 and aesthetic quality (AQ) scores above 50.00, removing samples with low visual fidelity, decoding artifacts and visual clutter. 
This filtering rule preserves 96.76\% of the collected clips.
\noindent\textbf{(3) Reference clarity verification.} 
The first frame of each clip is treated as the reference image. 
We apply Qwen3-VL-2B~\cite{bai2025qwen3vltechnicalreport} to automatically assess human subject visibility. 
The prompt explicitly specifies low visibility, back-facing poses, and missing facial cues as negative cases to filter out ambiguous or unclear references. 
After filtering, 81.45\% of the clips remain.
\noindent\textbf{(4) Dance video verification.} 
We apply Qwen3-VL-2B~\cite{bai2025qwen3vltechnicalreport} to perform automatic dance content classification. 
The prompt defines talking, posing, and static motions as negative conditions to filter out ambiguous clips. 
This filtering step retains 89.24\% of the clips.
\noindent\textbf{(5) Single-dancer filtering.} 
Directly detecting single-dancer videos is non-trivial. 
Instead, we identify and remove three representative negative scenarios using Qwen3-VL-2B~\cite{bai2025qwen3vltechnicalreport}: 
(i) group dance videos (e.g., duet or multi-person performances), 
(ii) follow-along imitation videos where a dancer replicates movements from an overlaid reference clip, and 
(iii) mirror-reflection artifacts commonly observed in studio settings, where reflections create the appearance of multiple performers. 
These three filtering criteria retain 80.35\%, 97.02\%, and 95.57\% of the clips, respectively, resulting in an overall retention rate of 74.45\%.
\noindent\textbf{(6) Scene stability filtering.} 
To address complex scene dynamics, we apply two Qwen3-VL-8B~\cite{bai2025qwen3vltechnicalreport} models to remove representative negative scenarios: 
(i) abrupt scene or subject transitions (e.g., sudden changes from one scene or subject to another), 
(ii) aggressive camera motion characterized by discontinuous or jittery movements. 
These two filtering criteria retain 93.95\% and 97.55\% of the clips, respectively, resulting in an overall retention rate of 91.65\%.

\begin{figure*}[t]
  \centering
  \includegraphics[width=\linewidth]{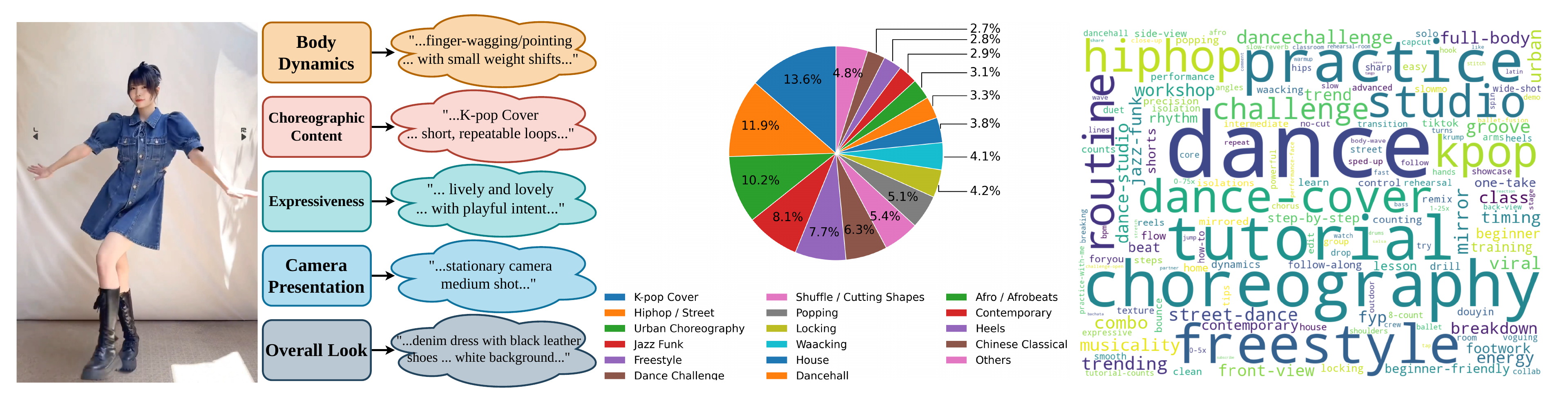}
  \caption{Dataset analysis. (Left) Display of typical annotations; (Middle) Dance genre distribution from a random sample of 400 entries; (Right) Word cloud summarizing the caption content.}
  \label{fig:annotation}   
\end{figure*}

\subsection{Data Annotation}
To better adapt the dataset to dance video generation, we adopt \textit{Choreography-Informed Text Annotations}. Fig.~\ref{fig:annotation} shows an example.
Under this scheme, each video is annotated by Qwen3-VL-8B~\cite{bai2025qwen3vltechnicalreport} from five independent yet complementary choreography-oriented aspects:
\noindent\textbf{(1) Body Dynamics.} 
Describes the concrete, moment-to-moment actions and body mechanics, including which body parts move and how (e.g., arm swings, hand waves, and hip isolations).
\noindent\textbf{(2) Choreographic Content.} 
Summarizes the \emph{dance-level semantics} of the choreography, including the dance genre/style (e.g., popping, hip-hop, jazz funk) and characteristic movement vocabulary or techniques (e.g., tutting, waving, Thomas flair).
\noindent\textbf{(3) Expressiveness.} 
Characterizes the overall emotional expression and performance intent conveyed by the dancer (e.g., joyful, playful, confident, melancholic), reflected through facial expression, posture, energy, and attitude.
\noindent\textbf{(4) Camera Presentation.} 
Describes shot composition, camera movement, and framing strategies that influence visual perception of the dance.
\noindent\textbf{(5) Overall Look.} 
Summarizes appearance attributes (e.g., clothing, styling) and environmental context that contribute to visual semantics.

\subsection{Data Statistic}
Finally, we present \textbf{CIPE-Dance}, a large-scale Internet-sourced dataset constructed via the Progressive Expert pipeline and equipped with Choreography-Informed annotations. 
To the best of our knowledge, \textbf{CIPE-Dance} is currently the largest open-source dataset for dance video generation, containing approximately 300K clips (over 400 hours of video content). See Fig. ~\ref{fig:annotation} for dataset analysis.
Beyond scale, CIPE-Dance provides comprehensive coverage across multiple orthogonal dimensions:
\noindent\textbf{(1) Genre Diversity.} 
The dataset spans over 30 dance genres, exhibiting a naturally long-tailed distribution that reflects real-world choreography trends.
\noindent\textbf{(2) Environmental Diversity.} 
Videos are collected from diverse real-world settings, including studios, stages, streets, and home environments, covering substantial variation in lighting and background complexity.
\noindent\textbf{(3) Performer Diversity.} 
CIPE-Dance features performers with diverse appearance attributes and performance styles, spanning professional stage recordings and casual social-media videos.
\noindent\textbf{(4) Motion Complexity.} 
Choreography ranges from rhythm-driven groove patterns to highly dynamic sequences involving large body displacement, jumps, and rapid limb articulation.
\noindent\textbf{(5) Camera Variability.} 
Clips include diverse viewpoints and framing styles (e.g., static, handheld, mild panning), while stability filtering ensures consistent identity and minimal abrupt transitions.
For evaluation, we randomly sample 100 clips from \textbf{CIPE-Dance} that are excluded from the training set to form a test split.

\begin{figure*}[t]
  \centering
  \includegraphics[width=\linewidth]{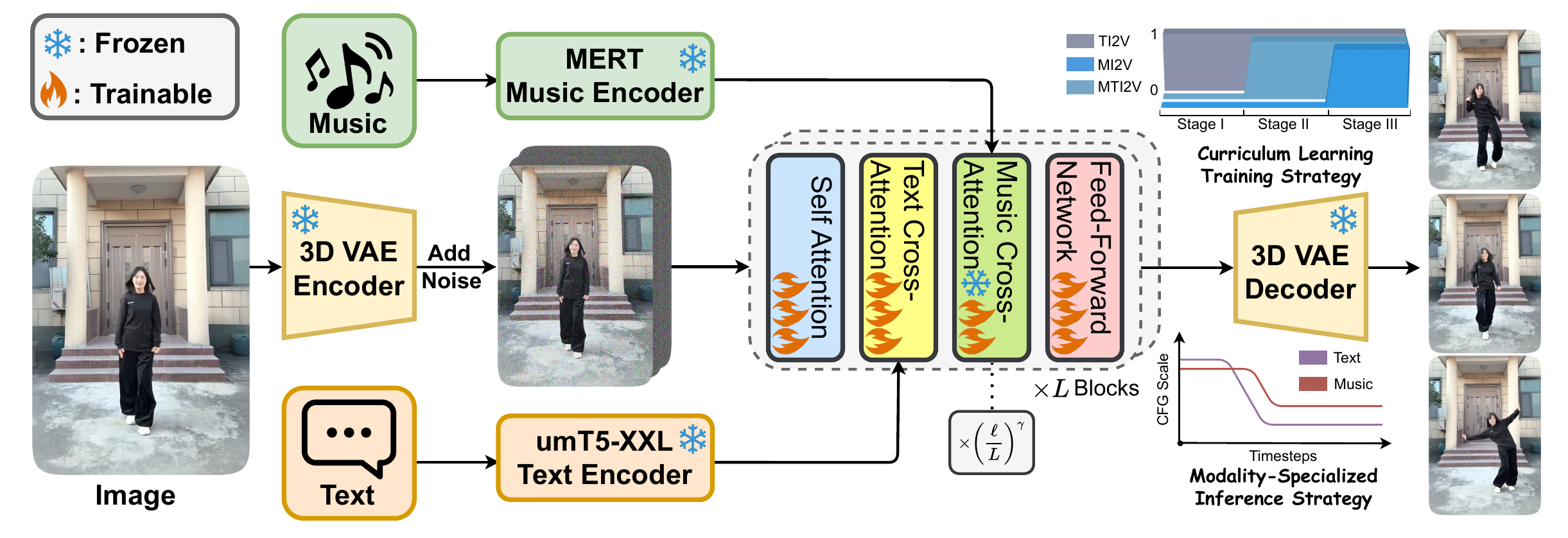}
  \caption{Overview of OmniDance: music–text progressive specialization model architecture, curriculum learning training strategy, and modality-specialized inference strategy.}
  \label{fig:overview}   
\end{figure*}

\section{Methodology}
\subsection{Overview}
To effectively integrate music conditioning into a Video Generation Foundation Model without degrading its original generative capabilities, \textbf{OmniDance} introduces a three-level design: 
(1) a music–text progressive specialization model architecture, 
(2) an easy-to-hard curriculum training strategy, and 
(3) a modality-specialized inference strategy. 
Moreover, \textbf{OmniDance} unifies TI2V, MI2V, and MTI2V generation within a single multimodal framework, as shown in Fig. ~\ref{fig:overview}.

\subsection{Model Architecture}
\subsubsection{Wan2.2-TI2V-5B}
Our backbone is Wan2.2-TI2V-5B, a Diffusion Transformer (DiT) pretrained foundation model for text/image-to-video generation. Each DiT block consists of a self-attention layer, a text-conditioned cross-attention layer, and a feed-forward network.
Wan2.2-TI2V-5B is equipped with a spatio-temporal VAE that downsamples the input by a factor of $(4,16,16)$ along the $(t,h,w)$ dimensions. This higher compression ratio reduces the latent resolution and provides approximately $4\times$ computational savings compared to prior configurations at the same output resolution.
Wan2.2-TI2V-5B is trained under the Flow Matching paradigm. Given the original latent representation $\mathbf{x}_0$ (data) and pure noise $\epsilon \sim \mathcal{N}(0, \mathbf{I})$, 
We define the linear optimal transport path $\mathbf{x}_t$ and its corresponding velocity field $\mathbf{v}_t$ as:
\begin{equation}
\mathbf{x}_t = (1 - t) \cdot \mathbf{x}_0 + t \cdot \epsilon, \quad \mathbf{v}_t = \epsilon - \mathbf{x}_0
\label{eq:flow_path}
\end{equation}
where the timestep \( t \) is randomly sampled from a uniform distribution \( \mathcal{U}(0,1) \). The velocity field predictor \( \mathbf{v}_\theta \), instantiated as WAN2.2-TI2V-5B, is trained to minimize:
\begin{equation}
\mathcal{L}_{\text{fm}} 
= \mathbb{E}_{\mathbf{x}_0, \epsilon, t}
\left[
\left\|
\mathbf{v}_\theta - \mathbf{v}_t
\right\|_2^2
\right].
\label{eq:stage1_loss}
\end{equation}

\subsubsection{Music-Text Progressive Specialization}
DiT-based foundation models naturally exhibit a coarse-to-fine learning hierarchy: early layers capture high-level and low-frequency global structures, while deeper layers progressively refine low-level and high-frequency details. 
In multimodal-driven dance video generation, music and text play complementary roles. Music primarily governs high-frequency temporal structures (e.g., rhythm, tempo, and energy evolution), whereas text specifies global semantic intent (e.g., motion style, choreography intent, and key pose constraints). Directly injecting both modalities into all layers may lead to representational competition rather than synergy.

To promote complementary specialization, we introduce \textbf{Music-Text Progressive Specialization}. Built upon WAN2.2-TI2V-5B, we augment each DiT block at layer $\ell \in \{1, \dots, L\}$ with an additional audio cross-attention module. The residual contribution of the audio branch is scaled in a depth-aware manner to progressively amplify music influence in deeper layers:
\begin{equation}
\mathbf{x}_\ell \leftarrow \mathbf{x}_\ell + \left( \frac{\ell}{L} \right)^{\gamma} \cdot \mathbf{r}_{\text{audio}}^\ell.
\label{eq:model_architecture}
\end{equation}
Here, $\mathbf{x}_\ell$ denotes the latent token features at layer $\ell$, and $\mathbf{r}_{\text{music}}^\ell$ is the residual update produced by the added music cross-attention branch. 
In early layers ($\ell \ll L$), text conditioning dominates, establishing the overall action layout and choreography structure. As generation proceeds to deeper layers ($\ell \rightarrow L$), the audio contribution gradually increases, enabling fine-grained refinement of motion dynamics, rhythmic alignment, and temporally localized movement details.

\subsection{Training Strategy}
Current mainstream video generation foundation models released in open-source form are primarily designed for text/image-to-video (TI2V) tasks and lack audio-related supervision signals. To effectively extend the TI2V capability of WAN2.2-5B toward unified TI2V, MI2V, and MTI2V generation, we adopt a \textit{Curriculum Training Strategy}, consisting of three easy-to-hard progressive stages.

\subsubsection{Stage I: Text-Adaptation Warm-up.}
In the first stage, \textbf{OmniDance} is trained exclusively on TI2V to adapt the pretrained foundation model to dance-specific textual descriptions. 
This phase stabilizes text-to-video alignment in the dance domain and ensures reliable semantic grounding before introducing additional modalities. 
The music branch remains frozen to prevent premature interference with the pretrained backbone. 
Formally, we optimize:
\begin{equation}
\min_{\theta_{\text{text}}}
\mathcal{L}_{\text{TI2V}},
\quad
\theta_{\text{music}} \text{ frozen}.
\end{equation}

\subsubsection{Stage II: Anchored Music Integration.}
In the second stage, a music branch is introduced while preserving the TI2V objective. 
The model is trained on both TI2V and MTI2V tasks, allowing music conditioning to be progressively learned under text guidance. 
This design mitigates abrupt degradation of the pretrained TI2V capability while establishing music–motion alignment in a controlled manner. 
To stabilize optimization when introducing the music branch, we zero-initialize its residual projection and apply a linear learning-rate warm-up to the newly introduced parameters, reducing the risk of disrupting the pretrained backbone during early training.
Formally, we train on:
\begin{equation}
\min_{\theta_{\text{text}}, \theta_{\text{music}}}
\mathcal{L}
\in
\{\mathcal{L}_{\text{TI2V}}, \mathcal{L}_{\text{MTI2V}}\}.
\end{equation}

\subsubsection{Stage III: Modality Decoupling and Specialization.}
In the final stage, \textbf{OmniDance} is trained on all three tasks—TI2V, MTI2V, and MI2V. 
Having acquired stable music–motion alignment in earlier stages, the model is encouraged to generate dance videos solely from music when textual input is absent. 
This stage promotes genuine music-driven specialization and reduces over-reliance on textual cues, enabling the model to capture intrinsic music–dance correlations. 
All parameters are optimized during this phase. 
Formally, we train on:
\begin{equation}
\min_{\theta_{\text{text}}, \theta_{\text{music}}}
\mathcal{L}
\in
\{\mathcal{L}_{\text{TI2V}}, 
\mathcal{L}_{\text{MTI2V}},
\mathcal{L}_{\text{MI2V}}\}.
\end{equation}

\subsection{Inference Strategy}
\noindent\textbf{Classifier-Free Guidance}
Under the Flow Matching paradigm, earlier timesteps ($\tau \to 0$) primarily model coarse, low-frequency global structures, while later timesteps ($\tau \to 1$) refine fine-grained, high-frequency details. 
Text conditioning typically conveys global semantic intent (e.g., choreography layout and motion style), whereas music provides temporally localized motion cues (e.g., rhythm, tempo, and energy variations). 
Motivated by this frequency alignment, we propose an \textit{Modality-Specialized CFG Strategy} that assigns text and music guidance to the timesteps where they are most effective. 
Specifically, text guidance is emphasized at early stages and gradually decays, while music guidance progressively increases toward later stages.
The unified guidance formulation is:
\begin{equation}
\begin{aligned}
\mathbf{v}_\theta^{\text{TI2V}}(x_\tau)
&=
\mathbf{v}_\theta(x_\tau, \varnothing, \varnothing)
+ \lambda_{t}(\tau)
\Bigl(
\mathbf{v}_\theta(x_\tau, text, \varnothing)
-
\mathbf{v}_\theta(x_\tau, \varnothing, \varnothing)
\Bigr), \\
\mathbf{v}_\theta^{\text{MI2V}}(x_\tau)
&=
\mathbf{v}_\theta(x_\tau, \varnothing, \varnothing)
+ \lambda_{m}(\tau)
\Bigl(
\mathbf{v}_\theta(x_\tau, \varnothing, music)
-
\mathbf{v}_\theta(x_\tau, \varnothing, \varnothing)
\Bigr), \\
\mathbf{v}_\theta^{\text{MTI2V}}(x_\tau)
&=
\mathbf{v}_\theta(x_\tau, \varnothing, \varnothing)
+ \lambda_{t}(\tau)
\Bigl(
\mathbf{v}_\theta(x_\tau, text, \varnothing)
-
\mathbf{v}_\theta(x_\tau, \varnothing, \varnothing)
\Bigr) \\
&\quad
+ \lambda_{m}(\tau)
\Bigl(
\mathbf{v}_\theta(x_\tau, text, music)
-
\mathbf{v}_\theta(x_\tau, text, \varnothing)
\Bigr).
\end{aligned}
\end{equation}
Over the generation trajectory, we use time-dependent guidance scales that both decay (text: $\lambda_t(\tau)$ from $5\!\to\!1$, music: $\lambda_m(\tau)$ from $4\!\to\!2$); since $\lambda_t$ decays more aggressively, the relative guidance gradually shifts from text-dominant to more music-influenced refinement.

\noindent\textbf{Long-Sequene Generation}
Following the WAN-style~\cite{wan2025wanopenadvancedlargescale} autoregressive extension, we adopt a sliding-window generation scheme: the last frame of the current clip is reused as the reference (starting) frame for the next window, enabling long-sequence synthesis by iteratively chaining short clips.

\begin{table*}[t]
\centering
\renewcommand{\arraystretch}{1.25}
\setlength{\tabcolsep}{4.5pt}
\caption{Quantitative comparison on the CIPE-Dance dataset for dance video generation task under different conditioning modalities, including Music-Image-to-Video (MI2V), Text-Image-to-Video (TI2V) and Music-Text-Image-to-Video(MTI2V).}
\label{tab:dance_video_comparison}
\resizebox{\linewidth}{!}{
\begin{tabular}{c|l|cccccc|cccc|cc}
\toprule
\textbf{Task} & \textbf{Method}
& \multicolumn{6}{c|}{\textbf{Video Quality}} 
& \multicolumn{4}{c|}{\textbf{Motion Quality}} 
& \multicolumn{2}{c}{\textbf{Alignment}} \\
\cmidrule(r){3-8} \cmidrule(r){9-12} \cmidrule(l){13-14}

&
& IQ$\uparrow$ & AQ$\uparrow$ & SC$\uparrow$ & BC$\uparrow$ & MS$\uparrow$ & TF$\uparrow$
& FID$_{k}$$\downarrow$ & FID$_{g}$$\downarrow$ & DIV$_{k}$$\uparrow$ & DIV$_{g}$$\uparrow$
& BAS$\uparrow$ & OC$\uparrow$ \\ 
\midrule

-- & Ground Truth 
& 67.11 & 54.16 & 91.51 & 93.62 & 97.23 & 95.36 
& -- & -- & 8.24 & 4.71 
& 0.288 & 12.86 \\

\midrule
\multirow{4}{*}{\rotatebox[origin=c]{90}{\textbf{MI2V}}}
& Hallo2~\cite{cui2024hallo2} [ICLR'25]
& 64.09 & 52.37 & 90.77 & 92.94 & 97.05 & 95.72
& 16.49 & 1.81 & 10.30 & 5.45
& 0.244 & 11.85 \\

& WAN-S2V~\cite{gao2025wan} [Tongyi'25]
& 65.33 & 55.81 & 89.97 & 91.92 & 97.34 & 95.47
& 21.47 & 2.83 & 10.25 & 5.39
& 0.283 & 12.10 \\

& Echomimic-V3~\cite{meng2025echomimicv3} [AAAI'26]
& 62.45 & 53.38 & 89.92 & 92.26 & 97.66 & 95.71
& 21.13 & 1.99 & 10.38 & 5.09 
& 0.274 & 12.02 \\

& \textbf{OmniDance (MI2V)}
& \textbf{65.61} & \textbf{54.96} & \textbf{92.63} & \textbf{93.80} & \textbf{98.09} & \textbf{96.65}
& \textbf{17.55} & \textbf{1.64} & \textbf{10.55} & \textbf{5.60} 
& \textbf{0.293} & \textbf{12.40} \\

\midrule
\multirow{3}{*}{\rotatebox[origin=c]{90}{\textbf{TI2V}}}
& CogVideoX1.5-5B~\cite{yang2024cogvideox} [ICLR'25]
& 55.77 & 51.58 & 87.32 & 91.50 & 97.05 & 95.66
& 15.51 & 2.82 & 10.36 & 5.55
& 0.222 & \textbf{12.53} \\

& HunyuanVideo~\cite{kong2025hunyuanvideosystematicframeworklarge} [Hunyuan'25]
& 63.92 & 54.11 & \textbf{90.70}
& 92.53 & 97.87 & 96.15
& 17.89 & 2.44 & 8.82 & 5.28
& 0.243 & 11.79 \\

& WAN2.2-TI2V-5B~\cite{wan2025wanopenadvancedlargescale} [Tongyi'25]
& 64.66 & 55.02 & 90.10 & 91.78 & 98.08 & 96.12
& 19.15 & 2.16 & 10.62 & 5.25
& 0.252 & 12.11 \\

& \textbf{OmniDance (TI2V)}
& \textbf{67.74} & \textbf{55.17} & 90.16 & \textbf{92.70} & \textbf{99.24} & \textbf{98.46}
& \textbf{13.55} & \textbf{2.04} & \textbf{11.15} & \textbf{5.79} 
& \textbf{0.259} & \textbf{12.53} \\

\midrule
& \textbf{OmniDance (MTI2V)}
&  \underline{67.77} &  \underline{55.76} &  \underline{94.34} &  \underline{94.54} & 99.22 &  \underline{98.49}
& 13.97 &  \underline{1.56} &  \underline{10.33} &  \underline{6.41} 
& 0.287 & 13.08 \\
\bottomrule
\end{tabular}
}
\end{table*}

\section{Experiment}
\subsection{Experimental Setup}
\subsubsection{Implementation Details}
We implement OmniDance using PyTorch on NVIDIA H20 GPUs. 
The backbone is Wan2.2-TI2V-5B with 30 DiT blocks. 
The audio encoder is MERT and the text encoder is umT5-XXL. 
Stage 1 trains for 5k steps, Stage 2 for 5k steps and Stage 3 for 10k, all with a batch size of 32, learning rate $1 \times 10^{-5}$, and the AdamW optimizer. 
For Music-Text Progressive Specialization, we set $\gamma = 1.5$ in Eq. \ref{eq:model_architecture}. 
We train OmniDance on 77-frame videos (5s at 16 FPS) at $704 \times 1280$ resolution using flow-matching sampling with 40 steps. 
The full training takes approximately 70 hours on 32 NVIDIA H20 GPUs, with about 80GB memory usage per GPU. During inference, generating a 5-second video on a single NVIDIA H20 GPU takes about 4.5 minutes for TI2V and MI2V, and about 7.2 minutes for MTI2V, with approximately 40GB GPU memory usage.

\subsubsection{Evaluation Metrics}
To comprehensively evaluate the performance of dance video generation task, we adopt metrics inspired by both VBench~\cite{li2024mvbench} and 3D Dance Generation~\cite{li2021ai}, assessing the results from three aspects: Video Quality, Motion Quality, and Alignment.
\textbf{(1) Video Quality.} Considering the dynamic and temporal nature of dance videos, we select several metrics from VBench~\cite{li2024mvbench}, including imaging quality (IQ), aesthetic quality (AQ), subject consistency (SC), background consistency (BC), motion smoothness (MS), and temporal flickering (TF).
\textbf{(2) Motion Quality.} We evaluate the diversity and fidelity of generated dance motions using DIV and FID computed on motion features. Specifically, we first extract 2D keypoint sequences using ViTPose~\cite{xu2022vitpose}, and then compute two types of motion features: (i) Kinetic Features~\cite{li2021ai} (k), which capture motion dynamics, and (ii) Geometric Features~\cite{li2021ai} (g), which encode spatial joint relationships.
\textbf{(3) Alignment.} To measure music–motion synchronization, we adopt the Beat Alignment Score (BAS)~\cite{li2021ai}. To assess text–video consistency, we use Overall Consistency (OC) from VBench~\cite{li2024mvbench}.
Additionally, BAS can also be applied to the TI2V task, as videos that align well with textual descriptions often exhibit rhythmic patterns similar to music. Conversely, OC can be used for the MI2V task, where the text prompt is simplified as “a person is dancing.”

\subsection{Comparison}
\subsubsection{Text-Image-to-Video}
We compare OmniDane (TI2V) against the representative strong TI2V foundation models, including CogVideoX1.5-5B~\cite{yang2024cogvideox} (ICLR'25), HunyuanVideo~\cite{kong2025hunyuanvideosystematicframeworklarge} (Hunyuan'25), and WAN2.2-TI2V-5B~\cite{wan2025wanopenadvancedlargescale} (Tongyi'25). 

Quantitatively (Tab.~\ref{tab:dance_video_comparison}), \textbf{OmniDance} demonstrates overall state-of-the-art performance in the TI2V task. Specifically, it achieves the best \emph{video quality} among TI2V baselines, with the highest IQ (67.74), AQ (55.17), BC (92.70), MS (99.24), and TF (98.46). It also yields better \emph{motion quality}, achieving the best fidelity (FID$_k$: 13.55, FID$_g$: 2.04) with strong diversity (DIV$_k$: 11.15, DIV$_g$: 5.79). Alignment is also improved, with BAS increasing to 0.259 (vs.\ 0.252 for WAN2.2) while maintaining competitive OC (12.53).

\begin{figure*}[t]
  \centering
  \includegraphics[width=\linewidth]{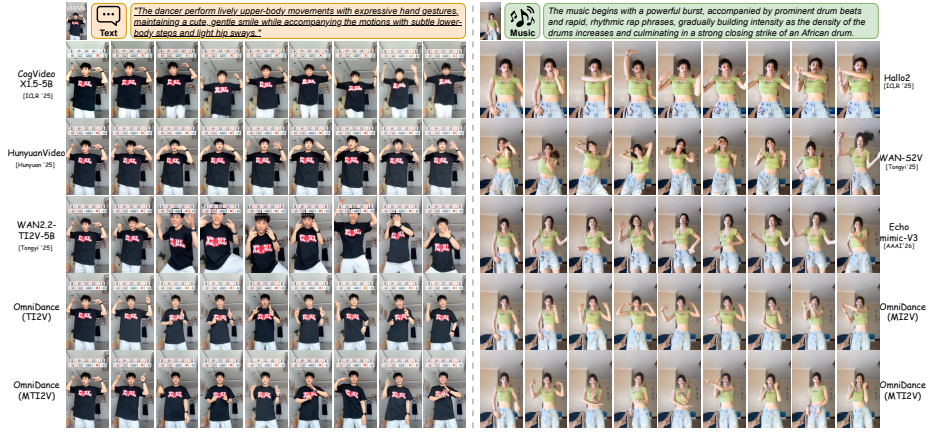}
  \caption{Comparison with SOTAs on Text-Image-to-Video (TI2V, left) and Music-Image-to-Video (MI2V, right) tasks on the CIPE-Dance dataset.}
  \label{fig:comparison}   
\end{figure*}

Qualitatively (Fig.~\ref{fig:comparison}), OmniDance generates more diverse hand articulations and maintains a natural smile, producing dance-like motion with subtle weight shifts and hip sways.

We attribute these gains to two factors: (i) the Stage-I \emph{text-adaptation warm-up} bridges the domain gap between generic TI2V pretraining and dance-specific semantics.  
(ii) CIPE-Dance’s high-quality, single-dancer, and scene-stable clips provide cleaner supervision for appearance fidelity and motion consistency.

\subsubsection{Music-Image-to-Video}
Due to limited open-source implementations specifically tailored for music-driven dance video generation, we compare OmniDance (MI2V) against two representative categories of audio-driven image animation baselines.
(i) Fine-tuning-based. We adapt Hallo2~\cite{cui2024hallo2} (ICLR'25) from talking-head animation to full-body dance by replacing its facial mask with a full-body mask.
(ii) Inference-only. We directly evaluate models advertised as general-purpose human motion generation systems without task-specific training, including WAN-S2V~\cite{gao2025wan} (Tongyi'25) and Echomimic-V3~\cite{meng2025echomimicv3} (AAAI'26).

Quantitatively (Tab.~\ref{tab:dance_video_comparison}), \textbf{OmniDance} exhibits the state-of-the-art performance in the MI2V task. Specifically, it achieves the best overall \emph{video quality} in terms of IQ (65.61), SC (92.63), BC (93.80), MS (98.09), and TF (96.65), while maintaining competitive AQ (54.96). It also improves \emph{motion quality}, achieving the best geometric fidelity (FID$_g$: 1.64) and the highest diversity (DIV$_k$: 10.55, DIV$_g$: 5.60). For \emph{alignment}, OmniDance reaches the best BAS (0.293) and OC (12.40), indicating stronger music-synchronized motion and overall consistency.

Qualitatively (Fig.~\ref{fig:comparison}), OmniDance produces rhythm-aware and intensity-progressive choreography with coherent gesture transitions that follow the drum beats, while maintaining stable identity and framing. In contrast, Hallo2 suffers from degraded visual quality with noticeable artifacts and blur; Echomimic-V3 often yields oversimplified and occasionally distorted motions; WAN-S2V tends to lose dancer identity over time (e.g., facial/appearance drift).

We attribute these gains to two factors: (i) the music branch is learned in a controlled curriculum (Stage II/III), enabling genuine MI2V generation without sacrificing stability; and (ii) the music-text progressive specialization together with modality-specialized CFG allocates music guidance to later refinement stages, improving beat-level synchronization.

\subsection{Multimodal Integration}
\noindent\textbf{Quantitative Analysis}
Tab.~\ref{tab:dance_video_comparison} shows that OmniDance performs strongly under both \emph{single-modality} settings. 
Under TI2V and MI2V, it achieves SOTA-level video and motion quality (e.g., TI2V: IQ 67.74 / MS 99.24; MI2V: SC 92.63 / BAS 0.293), indicating that enabling the music branch does not compromise the backbone’s core generation capability.
When both modalities are available, \textbf{OmniDance (MTI2V)} further strengthens overall multimodal consistency and joint control: SC and BC improve to 94.34 and 94.54, and OC increases to 13.08. 
Although a few metrics slightly decrease compared to the best single-modality counterpart (e.g., BAS 0.287 vs.\ 0.293 in MI2V, and DIV$_k$ 10.33 vs.\ 11.15 in TI2V), MTI2V yields a better overall trade-off by jointly preserving identity/scene stability and maintaining near ground-truth beat synchronization (0.287 vs.\ 0.288), demonstrating effective multimodal synergy.

\noindent\textbf{Qualitative Analysis}
As shown in Fig.~\ref{fig:comparison}, OmniDance benefits from clear cross-modal complementarity: adding music to TI2V enriches rhythmic structure and fine-grained motion dynamics, while adding text to MI2V strengthens semantic understanding. 
Overall, MTI2V leverages complementary cues from both modalities to generate more diverse and temporally coherent choreography, with richer expressions and more varied gesture sequences.

\noindent
We attribute the strong multimodal integration to:
(i) \emph{progressive specialization} that decouples text-guided semantics (shallow layers) from music-driven dynamics (deep layers);
(ii) an \emph{easy-to-hard curriculum} that learns music conditioning without degrading the pretrained TI2V capability; and
(iii) \emph{modality-specialized CFG} that emphasizes text early and music late along the generation trajectory.

\begin{figure*}[t]
\centering

\begin{minipage}{0.48\linewidth}
\centering
\captionof{table}{Ablation study on the model architecture on MTI2V task.}
\label{tab:3d_dance_comparison}
\renewcommand{\arraystretch}{1.5}
\scriptsize
\resizebox{0.96\linewidth}{!}{
\begin{tabular}{l|ccccc}
\toprule
 & AQ$\uparrow$ & SC$\uparrow$ & MS$\uparrow$ & BAS$\uparrow$ & OC$\uparrow$ \\

\midrule
Ground Truth          
& 54.16 & 91.51 & 97.23
& 0.288 & 12.86 \\

w/o. MTPS
& 54.18 & 90.61 & 93.74
& 0.257 & 12.53 \\

MTPS (R)
& 52.31 & 89.89 & 95.98
& 0.234 & 12.37 \\

w/o. MERT
& \textbf{56.10} & \textbf{95.72} & 98.64
& 0.241 & 12.78 \\

Full Model         
& \underline{55.76} & \underline{94.34} & \textbf{99.22}
& \textbf{0.287} & \textbf{13.08} \\

\bottomrule
\end{tabular}
}
\end{minipage}
\hfill
\begin{minipage}{0.48\linewidth}
\centering
\captionof{table}{Ablation study on the generative strategy on MTI2V task.}
\label{tab:3d_dance_comparison}
\renewcommand{\arraystretch}{1.5}
\scriptsize
\resizebox{0.96\linewidth}{!}{
\begin{tabular}{l|ccccc}
\toprule
 & AQ$\uparrow$ & SC$\uparrow$ & MS$\uparrow$ & BAS$\uparrow$ & OC$\uparrow$ \\

\midrule
Ground Truth          
& 54.16 & 91.51 & 97.23
& 0.288 & 12.86 \\

w/o. Stage I    
& 52.92 & 92.74 & 94.88
& 0.239 & 11.72 \\

w/o. Stage II   
& 51.38 & \textbf{94.61} & 98.11
& \textbf{0.293} & 12.71 \\

w/o. MS CFG
& 53.27 & 93.96 & 96.84
& 0.279 & \textbf{13.22} \\

Full Model  
& \textbf{55.76} & \underline{94.34} & \textbf{99.22}
& \underline{0.287} & \underline{13.08} \\

\bottomrule
\end{tabular}
}
\end{minipage}
\end{figure*}

\subsection{Ablation Study}
\subsubsection{Model Architecture}
We ablate key architectural components of OmniDance on the \textbf{MTI2V} task and report AQ/SC/MS/BAS/OC (Tab.~\ref{tab:3d_dance_comparison}). We evaluate: (i) \textbf{w/o.\ MTPS}, disabling the depth-aware audio residual scaling in Eq.~(\ref{eq:model_architecture}); (ii) \textbf{MTPS (R)}, reversing the schedule with stronger audio weights in shallow layers and weaker weights in deeper layers; (iii) \textbf{w/o.\ MERT}, replacing MERT with Wav2Vec2.0~\cite{baevski2020wav2vec}; and (iv) the \textbf{Full Model}. 

Specifically, \textbf{(1) Music-Text Progressive Specification.} Removing MTPS degrades temporal smoothness and music-motion synchronization (MS: 99.22 $\rightarrow$ 93.74; BAS: 0.287 $\rightarrow$ 0.257; OC: 13.08 $\rightarrow$ 12.53). Reversing the schedule further hurts performance (BAS: 0.234; SC: 89.89), supporting progressive strengthening of music in deeper layers; \textbf{(2) Music Representaion.} Replacing MERT with Wav2Vec2.0 improves appearance (AQ: 56.10; SC: 95.72) but reduces alignment and consistency (BAS: 0.241 vs.\ 0.287; OC: 12.78 vs.\ 13.08), indicating MERT provides more informative music cues for motion refinement.

\subsubsection{Generative Strategy}
We ablate the proposed training and inference strategy of OmniDance on the \textbf{MTI2V} task and report AQ/SC/MS/BAS/OC (Tab.~\ref{tab:3d_dance_comparison}). We evaluate: (i) \textbf{w/o.\ Stage I}, removing the Stage-I text-adaptation warm-up while keeping the remaining stages unchanged; (ii) \textbf{w/o.\ Stage II}, skipping the anchored music-integration stage and directly training the remaining stages; (iii) \textbf{w/o.\ MS CFG}, replacing our modality-specialized CFG with a unified scheme where \emph{text} and \emph{music} share the same guidance weight that decays from 5 to 3 along the flow; and (iv) the \textbf{Full Model}. 

Specifically, \textbf{(1) Curriculum stages.} Removing Stage I causes a clear performance drop across quality and alignment (AQ: 55.76 $\rightarrow$ 52.92; MS: 99.22 $\rightarrow$ 94.88; BAS: 0.287 $\rightarrow$ 0.239; OC: 13.08 $\rightarrow$ 11.72), indicating that text warm-up is critical for stabilizing dance-domain semantics before multimodal learning. Skipping Stage II yields strong beat alignment (BAS: 0.293) and competitive SC (94.61) but notably degrades appearance and consistency (AQ: 51.38; OC: 12.71), suggesting that abruptly introducing music can bias optimization toward synchronization at the expense of visual/text-consistent generation; \textbf{(2) Multimodality-Specific CFG.} Removing MS-CFG reduces smoothness and beat synchronization (MS: 99.22 $\rightarrow$ 96.84; BAS: 0.287 $\rightarrow$ 0.279), while slightly increasing OC (13.08 $\rightarrow$ 13.22), showing that modality-specification CFG strategy improves rhythm-driven refinement with a better overall trade-off.

\section{Conclusion}
In this work, we introduce \textbf{CIPE-Dance}, the largest dance video generation dataset to date, constructed via a progressive expert-based pipeline and equipped with choreography-informed text annotations.
Next, we propose \textbf{OmniDance}, a unified framework that supports multimodal driven dance video generation, enabled by a music-text progressive specialization model architecture, an easy-to-hard curriculum training strategy, and a modality specialized inference strategy.
Extensive experiments on \textbf{CIPE-Dance} demonstrate that \textbf{OmniDance} achieves state-of-the-art performance, while exhibiting strong multimodal integration capability.
Although OmniDance does not yet support real-time generation, real-time generation plays an important role for this field. We believe that combining \textit{Long-Video Generation via Distillation} offers a promising pathway toward efficient dance video synthesis.

\section{Acknowledge}
This work was supported in part by the National Natural Science Foundation of China under Grants 62436010, 72572090, 62572474 and 62172421.


%
%
\bibliographystyle{splncs04}
\bibliography{main}
\end{document}